\documentclass{article}
\usepackage{ijcai26}
\usepackage{threeparttable}
\usepackage{times}
\usepackage{placeins}

\usepackage{soul}
\usepackage{url}
\usepackage[hidelinks]{hyperref}
\usepackage[utf8]{inputenc}
\usepackage[small]{caption}
\usepackage{graphicx}
\usepackage{multirow}
\usepackage{amssymb}
\usepackage{amsmath}
\usepackage{amsthm}
\usepackage{booktabs}
\usepackage{algorithm}
\usepackage{algorithmic}
\usepackage[switch]{lineno}

\title{Meta-Soft: Leveraging Composable Meta-Tokens for Context-Preserving KV Cache Compression }

\author{
Wei Luo$^{1,2}$
\and
Yi Huang$^3$
\and
Songchen Ma$^4$
\and
Huanyu Qu$^{1,2}$
\and
Jiang Cai$^1$
\and
Mingkun Xu$^{1}$\thanks{Corresponding author.}
\\
\affiliations
$^1$Guangdong Institute of Intelligence Science and Technology\\
$^2$University of Macau\\
$^3$Chinese University of Hong Kong, Shenzhen\\
$^4$Hong Kong University of Science and Technology\\
\emails
\{luowei, quhuanyu, caijiang, xumingkun\}@gdiist.cn,
225040495@link.cuhk.edu.cn,
songchenma@ust.hk
}

\begin{document}

\maketitle

\begin{abstract}
The KV cache used in large language models has linearly growing time complexity, so LLMs face memory blow-up and reduced decoding efficiency when they process long contexts. Current KV Cache eviction has become an important research direction; however, existing methods based on fixed Soft Tokens (e.g., Judge Q) rely on a static parameter set as the query to evaluate the importance of KV pairs, so they cannot adapt dynamically to different input prompts, and they cannot precisely capture complex and changing task relevance. Also, evicted KV pairs are discarded permanently, so this causes irreversible information loss and context breaks. To address this problem, we propose Meta-Soft, a dynamic compression framework based on probe-driven context integration.
Specifically, we build a meta-library with a learnable orthogonal basis matrix $\mathcal{L}$, and we use a selector network with Gumbel-Softmax to produce differentiable sparse combination weights, so we dynamically synthesize the most targeted $k$ Soft Tokens from the input prompt features. We append these Soft Tokens to the end of the input sequence to probe key information. We also introduce an attention-flow based integration mechanism, which redistributes the semantic information of removed tokens into retained tokens, and this keeps the dropped context information effectively. Experiments on multiple datasets show that our method outperforms existing state-of-the-art eviction methods and provides a new solution for KV Cache compression.
    
\end{abstract}

\section{Introduction}

In recent years, large language models (LLMs) have made revolutionary progress in natural language processing, and from GPT-3.5 to Llama-3, these models have shown strong ability in long-text understanding, complex reasoning, and multi-turn dialogue.
LLM inference usually uses an autoregressive mechanism, and in order to avoid recomputing the Key--Value pairs of historical tokens at each generation step, LLMs introduce the KV Cache mechanism \cite{pope2023}.
However, as the input sequence length increases, the GPU memory usage of the KV Cache grows linearly.
When LLMs handle long-context tasks such as ``needle-in-a-haystack'' retrieval or long-document summarization, the huge KV Cache not only causes GPU memory overflow, but it also significantly reduces decoding throughput \cite{flashattention2022,vllm2023}.
So, how to compress the KV Cache efficiently while maintaining model performance has become a key challenge in current LLM deployment.

To address this problem, the research community has proposed multiple directions for KV Cache compression, and many of the latest studies focus on KV Cache eviction strategies.
The core assumption of this strategy is that not all historical KV pairs are equally important for current generation.
We can reduce GPU memory usage significantly by identifying and keeping ``important'' KV pairs.
However, existing studies still have clear limitations in two key steps, namely importance evaluation and the handling of evicted KV pairs.

Most existing eviction methods use attention weights to measure the importance of KV pairs.
Classic methods such as H2O \cite{h2o2023} and StreamingLLM \cite{streamingllm2024} mainly rely on accumulated attention scores or the ``Attention Sink'' phenomenon, and they keep KV pairs that are early in position or have high accumulated weights.
However, these methods usually use the current decoding window as the query to compute the importance of historical KV pairs.
As RoCo \cite{roco2024} and SnapKV \cite{snapkv2024} point out, this local greedy strategy cannot capture global dependencies, so it fails in scenarios that require long-range backtracking, and it is also far from the true global query distribution during decoding.
To address this issue, Lookahead Q-Cache \cite{lookahead2024} introduces a pseudo-query mechanism, and it tries to make more consistent eviction decisions by predicting the future query distribution, which mitigates the local short-sightedness problem.
Further, Judge Q \cite{judgeq2025} proposes to train a set of specific ``judge tokens'' to learn an optimal retention policy, and it achieves a clear improvement over heuristic rules.
However, Judge Q still relies on a static set of learnable parameters.
We argue that fixed parameters cannot adapt to very different task patterns at the same time.
This lack of dynamic adaptation to different input prompts makes existing methods hard to capture complex and changing long-context relevance precisely.

\begin{figure}[t]
 \centering
 \includegraphics[width=\columnwidth]{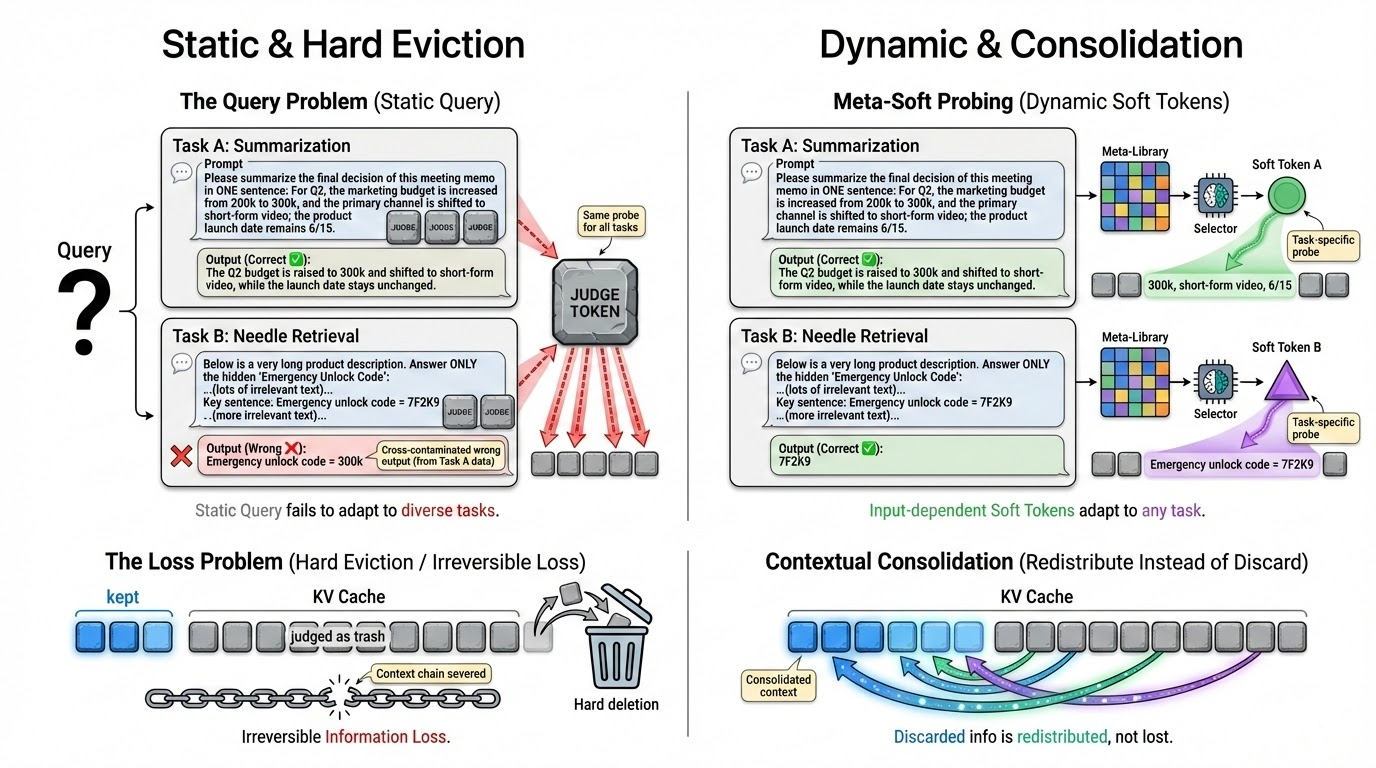}
 \caption{Motivation and overview of Meta-Soft. \textbf{Left:} Existing KV-cache compression often relies on \emph{static queries} for eviction, which fail to adapt across diverse tasks and may cause cross-task mismatch; moreover, \emph{hard eviction} permanently deletes KV entries, leading to irreversible information loss and broken context. \textbf{Right:} Meta-Soft uses \emph{input-dependent dynamic soft tokens} synthesized from a meta-library to probe task-relevant salient tokens, and performs \emph{contextual consolidation} that redistributes the information of discarded tokens into retained ones instead of deleting them, improving robustness under long-context compression.}

 \label{fig:motivation}
\end{figure}

Besides importance evaluation for KV pairs, how to handle KV pairs that are judged as ``not important'' is also a major challenge.
Most mainstream methods (e.g., Scissorhands \cite{scissorhands2023}, FastGen \cite{fastgen2024}) adopt a Top-$K$ hard eviction strategy: once the score is below a threshold, the KV pair is discarded permanently.
This ``keep-or-drop'' operation causes irreversible information loss and context breaks, and it can lead to hallucination \cite{mikv2024}.
Although some recent KV merging works, such as CaM \cite{cam2024} and ZipCache \cite{zipcache2024}, try to reduce the number of KV pairs by averaging or clustering similar KV pairs, this naive merging often causes semantic mixing, so accuracy drops when the model needs precise queries.As shown in Figure~\ref{fig:motivation}, existing compression strategies do not achieve a balance between dynamic adaptation and information preservation.

To address the above problems of poor adaptation in static-query methods and information loss caused by hard eviction or simple merging, we propose a new dynamic compression framework, Meta-Soft.
First, we design and build a Meta-Library that contains an orthogonal basis matrix, and we use a selector network to synthesize Soft Tokens that best match the current task, based on the global semantic features of the current prompt, while using Gumbel-Softmax to produce dynamic synthesis.
These probes are appended to the end of the input sequence at the embedding layer, and through one forward attention computation, they act as queries to find global KV pairs that are truly important for the current task.
Second, we propose a context integration mechanism.
Unlike hard eviction that directly discards, or direct merging that blindly averages, we use attention flow to compute semantic similarity between evicted tokens and retained tokens.
We keep the semantic information of an evicted token by moving it into the retained token that is most similar to it.
This mechanism keeps the clean semantics of core tokens, and it also recovers information from evicted context effectively, so it achieves efficient and lossless compression.

Our main contributions are as follows:
\begin{itemize}
  \item We propose a Soft Token generation mechanism based on a meta-library and dynamically synthesized weights, and it overcomes the limitations of existing static-query methods, so it supports adaptive perception for prompts from different tasks.
  \item We propose, for the first time, an attention-flow information compensation strategy, and it addresses information breaks caused by hard eviction and semantic blur caused by naive KV merging.
\end{itemize}

\section{Related Work}
\subsection{KV Cache Eviction and Merging}
KV cache compression has become a key research frontier for alleviating the memory bottleneck of long-context large language models (LLMs). Early \textbf{Hard Eviction} methods mainly relied on static heuristic strategies to identify and discard redundant tokens. \textit{Scissorhands}~\cite{scissorhands2023} first leveraged the “importance persistence” assumption; accordingly, it maintains a fixed-size cache by evicting tokens with low accumulated attention scores. Building on this, \textit{H2O}~\cite{h2o2023} identifies “Heavy Hitters”---namely, a small set of core tokens---and then adopts a greedy eviction strategy based on cumulative scores. To stabilize long-sequence generation, \textit{StreamingLLM}~\cite{streamingllm2024} discovers the “attention sink” phenomenon, and thus retains the initial tokens as well as the most recent tokens. In response to the static limitations of prior work, \textit{FastGen}~\cite{fastgen2024} introduces an adaptive attention strategy, dynamically adjusting the eviction budget per attention head. Furthermore, \textit{SnapKV}~\cite{snapkv2024} optimizes the prefilling stage by observing that attention heads tend to concentrate on specific clusters. However, these methods often fall into “local myopia” because they only rely on the recent window to evaluate importance. To address this issue, \textit{Lookahead Q-Cache}~\cite{lookahead2024} uses \textit{Pseudo Queries} to predict future attention distributions, thereby enabling more consistent eviction decisions. Meanwhile, \textit{D${2}$O}~\cite{d2o2024} further improves this discrimination process via dynamic decision operations, efficiently pruning irrelevant context. More recently, \textit{Judge Q}~\cite{judgeq2025} proposes \textit{Trainable Queries} to learn optimal information-retention representations, which represents the current state of the art; nevertheless, it still lacks dynamic adaptability to different input prompts.

In parallel with eviction, \textbf{Retention and Merging} strategies aim to integrate information rather than simply discarding it. \textit{ToMe}~\cite{tome2023} introduces token merging via bipartite matching, using feature similarity to shorten sequence length. Then, \textit{AutoCompressors}~\cite{autocomp2023} extends this by training summary tokens to compress context, but it requires expensive fine-tuning. \textit{CaM}~\cite{cam2024} proposes \textit{Cache Merging}, identifying and merging redundant KV pairs across attention heads and layers to preserve semantic integrity. On top of that, \textit{ZipCache}~\cite{zipcache2024} reduces aliasing effects during merging through channel normalization, improving efficiency. Moreover, \textit{PyramidKV}~\cite{pyramidkv2024} observes differences in capacity demand across layers and proposes a pyramid structure, performing more aggressive merging in deeper layers. \textit{Gist Tokens}~\cite{gist2024} uses dedicated virtual tokens to compress long prompts into compact activation vectors. Furthermore, \textit{Context Compression}~\cite{lo-co2024} employs low-rank approximation to map historical context into fixed latent states. Finally, \textit{ZeroMerge}~\cite{zeromerge2024} achieves parameter-agnostic state-of-the-art performance via KV-pair merging without additional training, yet it still faces the “semantic aliasing” problem---that is, semantic confusion caused by blind merging.



\section{Methodology}

\subsection{Method Overview}
\begin{figure*}[t]
    \centering
    \includegraphics[width=\textwidth]{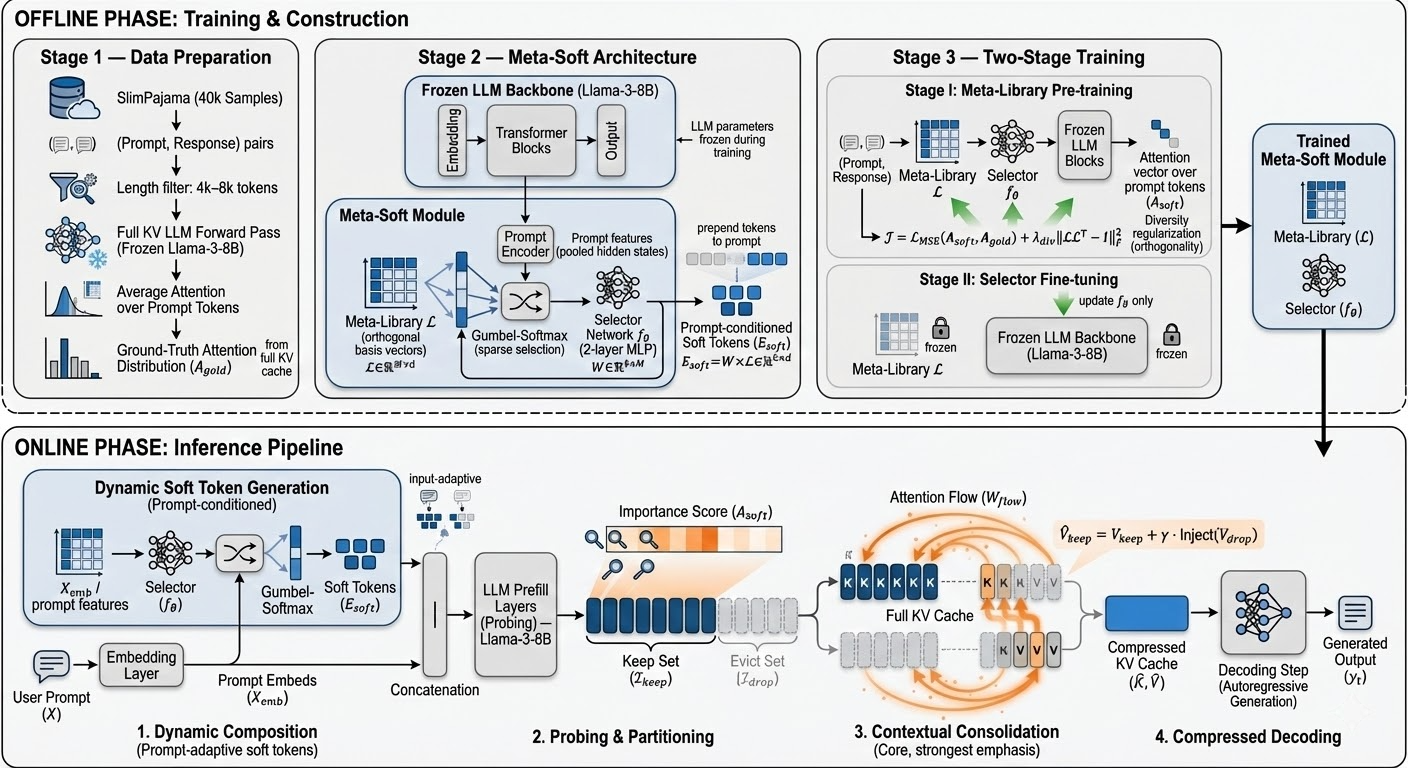}
    \caption{\textbf{Meta-Soft framework overview.} Meta-Soft trains a Meta-Library and selector offline with Ground-Truth Attention supervision and compresses the KV cache online by generating prompt-conditioned soft tokens to probe, partition, and consolidate context for decoding.}

    \label{fig:metasoft-framework}
\end{figure*}

We propose \textbf{Meta-Soft}, a dynamic framework for KV cache compression that leverages input-adaptive probing and semantic consolidation. As illustrated in Fig.~\ref{fig:metasoft-framework}, the overall workflow consists of an \emph{offline} preparation phase and an \emph{online} inference phase. In the offline stage, we optimize a \textbf{Meta-Library} and a lightweight selector through a two-stage training paradigm, where the supervision signal is the \textbf{Ground-Truth Attention} distribution $A_{gold}$ extracted from a frozen LLM backbone. In the inference stage, the module \emph{dynamically} synthesizes prompt-specific \textbf{Soft Tokens} ($E_{soft}$) conditioned on the input prompt; these soft tokens are concatenated to the embedding sequence to probe the full-context KV cache. Based on the probing-derived importance scores, we partition the cache into retained and evicted sets and execute \textbf{Contextual Consolidation}, where evicted semantic information is redistributed into the retained set via an attention-flow mechanism, enabling high-fidelity context preservation without additional LLM training.

\subsection{Problem Formulation}
Let $X = [x_1, \dots, x_L]$ be the input prompt with length $L$. Let $X_{emb} \in \mathbb{R}^{L \times d}$ denote the corresponding embeddings, where $d$ is the hidden dimension. In a Transformer layer with $H$ attention heads and head dimension $d_k$, $X_{emb}$ is projected into Key and Value matrices $K, V \in \mathbb{R}^{L \times (H \cdot d_k)}$. Our objective is to determine a compressed subset of indices $\mathcal{I}_{keep} \subset \{1, \dots, L\}$ such that $|\mathcal{I}_{keep}| = B$, where $B$ is the compression budget. 
Rather than simply discarding the evicted set $\mathcal{I}_{drop} = \{1, \dots, L\} \setminus \mathcal{I}_{keep}$, we construct an augmented cache $(\hat{K}_{keep}, \hat{V}_{keep})$ to satisfy $P(y | \hat{K}_{keep}, \hat{V}_{keep}) \approx P(y | K, V)$, where $y$ is the generated response.

\subsection{Data Acquisition and Meta-Library Construction}
\paragraph{Data Acquisition}
We utilize \textbf{SlimPajama}~\cite{shen2024slimpajamadcunderstandingdatacombinations} to curate a training set of \textbf{40,000 samples}, each containing a prompt $x_{prompt}$ and a response $x_{response}$. We extract the \textbf{Ground-Truth Attention Distribution} $A_{gold} \in \mathbb{R}^L$ as the supervision signal. Let $\text{Attn}^{(h)}_{i,j}$ be the attention weight of the $i$-th response token toward the $j$-th prompt token in head $h$. $A_{gold}$ is defined as:
\begin{equation}
    A_{gold, j} = \frac{1}{H \cdot L_{res}} \sum_{h=1}^H \sum_{i=1}^{L_{res}} \text{Attn}^{(h)}_{i, j}, \quad \forall j \in \{1, \dots, L\}
\end{equation}

\paragraph{Two-Stage Training Strategy}
The Meta-Library $\mathcal{L} \in \mathbb{R}^{M \times d}$ and the selector $f_\theta$ are optimized through a coordinated two-stage process. The total loss function is defined as:
\begin{equation}
    \mathcal{J} = \mathcal{L}_{MSE}(A_{soft}, A_{gold}) + \lambda_{div} || \mathcal{L}\mathcal{L}^T - I ||_F^2
\end{equation}
The first term, $\mathcal{L}_{MSE}$, ensures the synthesized probes accurately mimic the Ground-Truth's attention behavior, while the second term serves as an orthogonality regularization. This diversity constraint forces the basis vectors in $\mathcal{L}$ to span a wider representation space, preventing feature redundancy and ensuring that the meta-library can capture multifaceted semantic dependencies.

\begin{itemize}
    \item \textbf{Stage I: Joint Optimization.} We jointly update $\mathcal{L}$ and $f_\theta$ to establish the foundation of the meta-basis. In this stage, the gradients flow through the Gumbel-Softmax composition to both the library and the selector, allowing the library to learn a set of generic, orthogonal semantic atoms that represent common attention patterns across the 40,000 samples.
    \item \textbf{Stage II: Selector Fine-tuning.} We freeze the optimized Meta-Library $\mathcal{L}$ and the LLM backbone, fine-tuning only the selector $f_\theta$. This stage focuses on refining the task-specific combination strategy. By fixing the basis vectors, we prevent representation drift and force the selector to learn the optimal mapping from diverse prompt features to the established semantic space.
\end{itemize}

\subsection{Soft Token Probing and Cache Partitioning}
During prefill, $E_{soft}$ is concatenated to the prompt embeddings: 
\begin{equation}
    X_{input} = [X_{emb}; E_{soft}] \in \mathbb{R}^{(L+k) \times d}
\end{equation}
where $X_{emb}$ represents the initial prompt embeddings and $E_{soft}$ denotes the $k$ synthesized soft tokens.

\paragraph{Attention Probing}
Let $W_Q, W_K \in \mathbb{R}^{d \times (H \cdot d_k)}$ be projection matrices. The probe queries and prompt keys are $Q_{probe} = E_{soft} W_Q$ and $K_{prompt} = X_{emb} W_K$. The probing score $S \in \mathbb{R}^{k \times L}$ is:
\begin{equation}
    S = \frac{Q_{probe} (K_{prompt})^T}{\sqrt{d_k}}
\end{equation}
The predicted distribution $A_{soft} \in \mathbb{R}^L$ is obtained by:
\begin{equation}
    \alpha_i = \text{Softmax}(S_{i, :}), \quad A_{soft} = \frac{1}{k} \sum_{i=1}^k \alpha_i
\end{equation}

\paragraph{Cache Partitioning}
Based on $A_{soft}$ and budget $B$, we partition the cache:
\begin{equation}
    \mathcal{I}_{keep} = \text{TopK}(A_{soft}, B), \quad \mathcal{I}_{drop} = \{1, \dots, L\} \setminus \mathcal{I}_{keep}
\end{equation}
This yields the partitioned tensors $(K_{keep}, V_{keep})$ for the retained set and $(K_{drop}, V_{drop})$ for the evicted set. This partitioning constitutes the foundation for the subsequent contextual consolidation stage, ensuring that the KV pairs of the most semantically salient tokens are explicitly preserved, while the remaining tokens are prepared for information aggregation.

\subsection{Probing-Driven Contextual Consolidation}
We do not directly discard $\mathcal{I}_{drop}$; instead, we utilize \textbf{Attention Flow} to inject its information into $\mathcal{I}_{keep}$. 

\paragraph{Diversity-Preserving Attention Flow Aggregation}
We compute the flow weights by measuring the key-space similarity Directly using $\mathrm{Softmax}(K_{drop}K_{keep}^\top)$ as key-space similarity. may cause many evicted tokens to collapse onto a few highly similar kept tokens, leading to overwriting and information loss. 
To mitigate this, we adopt a simple \emph{load-balanced sparse routing} scheme that explicitly penalizes overloaded kept tokens while keeping the routing definition lightweight and efficient.

We compute key-space similarities:
\begin{equation}
S_{sim}=\frac{K_{drop}K_{keep}^\top}{\sqrt{d_k}}.
\end{equation}
For each dropped token $i$, let $\mathcal{N}_m(i)$ be the indices of its top-$m$ largest entries in $S_{sim}(i,\cdot)$. 
We define a sparse row-stochastic assignment:
\begin{equation}
A_{ij}=
\begin{cases}
\frac{\exp(S_{sim,ij}/\tau)}{\sum\limits_{j'\in\mathcal{N}_m(i)}\exp(S_{sim,ij'}/\tau)}, & j\in\mathcal{N}_m(i),\\[6pt]
0, & \text{otherwise},
\end{cases}
\end{equation}
where $\tau$ controls the sharpness and $m\ll |\mathcal{I}_{keep}|$ enforces sparsity.

To prevent a few kept tokens from absorbing excessive total mass, we compute the total attention mass assigned to it of each kept token:
\begin{equation}
\ell_j=\sum_{i\in \mathcal{I}_{drop}} A_{ij},
\end{equation}
and apply a column reweighting factor that down-weights overloaded columns:
\begin{equation}
b_j=\frac{1}{\ell_j+\epsilon},\qquad 
\tilde{W}_{ij}=A_{ij}\,b_j,
\end{equation}
then renormalize each row to ensure that every dropped token’s outgoing weights still sum to 1.
\begin{equation}
W_{flow,ij}=\frac{\tilde{W}_{ij}}{\sum_{j'}\tilde{W}_{ij'}}.
\end{equation}
Finally, we aggregate evicted values into the kept set:
\begin{equation}
\Delta V = W_{flow}^\top V_{drop},
\end{equation}
and update the kept values with a load-adaptive gate to avoid destructive overwrite:
\begin{equation}
g_j=\mathrm{clip}\!\left(\frac{\alpha}{\ell_j+\epsilon},\,0,\,1\right),\quad
\hat{V}_{keep,j}=V_{keep,j}+\gamma\,(g_j\,\Delta V_j),
\end{equation}
where $\alpha=\frac{|\mathcal{I}_{drop}|}{|\mathcal{I}_{keep}|}$ is the target average load, $\gamma$ is the consolidation coefficient, and $K_{keep}$ remains unchanged to preserve positional information of KV pairs.

\subsection{Integration into Inference Pipeline}
The integration of Meta-Soft is seamless  for the LLM backbone. The entire process—Soft Token generation, Probing, and Contextual Consolidation—occurs solely during the \textbf{Prefill Phase}.
Once the compressed cache $(\hat{K}_{keep}, \hat{V}_{keep})$ is constructed, the soft tokens $E_{soft}$ and the evicted set are removed from memory. The subsequent \textbf{Decoding Phase} proceeds using standard attention mechanisms over the compressed cache, incurring no additional computational overhead per generation step.
\section{Experiments}

\subsection{Experimental Setup}
\paragraph{Models and Datasets}
We evaluate Meta-Soft on long-context KV cache compression using Llama-3.1-8B-Instruct (from the Llama 3 family)~\cite{llama3herd2024} and Mistral-7B-Instruct-v0.3~\cite{mistral7b2023}.
Our experiments cover PG19~\cite{rae2019compressive}, OpenWebText2~\cite{openwebtext2}, and long-context benchmarks including LongBench~\cite{longbench2023} and RULER~\cite{ruler2024}.

\paragraph{Baselines}
We compare Meta-Soft against several state-of-the-art KV cache compression methods, including H2O~\cite{h2o2023}, SnapKV~\cite{snapkv2024}, StreamingLLM~\cite{streamingllm2024}, LAQ (Lookahead Q-Cache)~\cite{lookahead2024}, Judge Q~\cite{judgeq2025}, CaM~\cite{cam2024}, ZeroMerge~\cite{zeromerge2024}, and AnDPro~\cite{geng2025andpro}.We additionally report the performance with a full KV cache as an upper bound.

\paragraph{Implementation Details}
Our Meta-Library is initialized with $M=512$ basis vectors, and we synthesize $k=32$ soft tokens for probing by default.
Training is conducted on three NVIDIA A100 (80GB) GPUs.
We first perform joint training on a 40k-sample subset of SlimPajama, where gradients flow through the Gumbel-Softmax composition to update both the Meta-Library $\mathcal{L}$ and the selector $f_\theta$, for 5 epochs.
We then freeze $\mathcal{L}$ and the LLM backbone and fine-tune only the selector $f_\theta$ on 10,000 samples from ShareGPT for 3 epochs~\cite{vicuna2023}.The overall convergence time is approximately 5.5 hours.
We use the AdamW optimizer with a learning rate of $1\times 10^{-4}$, and anneal the Gumbel-Softmax temperature $\tau$ from 1.0 to 0.1 (aligned with the training steps).

\subsection{Language Modeling Evaluation}
We evaluate Meta-Soft with Llama-3.1-8B-Instruct on PG19 and OpenWebText2 to test whether it preserves language coherence under KV cache compression. Following the table setting, we report results at context lengths of 4k and 16k, and compare the Full KV setting (no compression) with a single compressed cache size of $B=256$. For each dataset and each configuration, we randomly sample 1,000 examples and report the mean perplexity (PPL). As shown in Table~\ref{tab:ppl}, Meta-Soft consistently achieves the lowest PPL among H2O, SnapKV, Judge Q, and ZeroMerge on both PG19 and OpenWebText2, indicating improved fidelity under KV cache compression.

\begin{table}[t]
\centering
\begingroup
\resizebox{\columnwidth}{!}{%
\setlength{\tabcolsep}{5pt} 
\renewcommand{\arraystretch}{1.0}
\begin{tabular}{llcc}
\toprule
Dataset & Method & context length= 4k & context length=16k \\
\midrule
\multirow{6}{*}{PG19}
& Full KV    & 6.81 & 7.23 \\
& H2O        & 7.27 & 7.79 \\
& SnapKV     & 7.24 & 7.72 \\
& Judge Q    & 7.11 & 7.58 \\
& ZeroMerge  & 7.13 & 7.63 \\
& Meta-Soft  & 7.05 & 7.49 \\
\midrule
\multirow{6}{*}{OpenWebText2}
& Full KV    & 5.45 & 5.72 \\
& H2O        & 5.82 & 6.12 \\
& SnapKV     & 5.83 & 6.09 \\
& Judge Q    & 5.71 & 5.94 \\
& ZeroMerge  & 5.75 & 5.98 \\
& Meta-Soft  & 5.68 & 5.87 \\
\bottomrule
\end{tabular}%
}
\endgroup
\caption{Mean PPL (lower is better) of Llama-3.1-8B-Instruct on PG19 and OpenWebText2 at context lengths 4k/16k, comparing Full KV to KV cache compression ($B=256$).}
\label{tab:ppl}
\end{table}

\subsection{Results on LongBench}
Evaluation on \textbf{LongBench} provides a holistic view of the model's capability in real-world long-context tasks. Table~\ref{tab:main-longbench} reports the LongBench results under compact KV caches ($B\in\{128,256\}$). Meta-Soft achieves the best average score for both backbones, outperforming SnapKV by $1.7$--$3.2$ points and Judge Q by $0.9$--$1.1$ points across cache sizes, with gains observed in most task categories. Moreover, Meta-Soft preserves $92.9$--$97.2\%$ of the Full-KV average performance, indicating strong long-context retention under limited KV budgets.

\begin{table*}[t]
\centering
\begingroup
\setlength{\tabcolsep}{9.0pt}
\renewcommand{\arraystretch}{0.92}
\small
\resizebox{\textwidth}{!}{%
\begin{tabular}{lccccccccccccccccc}
\toprule
\textbf{Method} &
\multicolumn{3}{c}{Single-Doc QA} &
\multicolumn{3}{c}{Multi-Doc QA} &
\multicolumn{3}{c}{Summ.} &
\multicolumn{3}{c}{Few-shot} &
\multicolumn{1}{c}{Synth.} &
\multicolumn{3}{c}{Code} &
\multicolumn{1}{c}{Avg.} \\
\cmidrule(lr){2-4}\cmidrule(lr){5-7}\cmidrule(lr){8-10}\cmidrule(lr){11-13}\cmidrule(lr){14-14}\cmidrule(lr){15-17}\cmidrule(lr){18-18}
& NQA & QSP & MF & HotpotQA & 2Wiki & Musq & GovR & QM & MN & TREC & Trivia & SAM & PGM & Pre & Lcc & RBP & Avg. \\
\midrule

\multicolumn{18}{c}{\textbf{Llama-3.1-8B-Instruct}} \\
\midrule

Full KV
& 30.62 & 46.45 & 57.34 & 58.21 & 50.12 & 32.42 & 35.15 & 25.54 & 27.91 & 73.42 & 92.68 & 43.12 & 8.42 & 100.00 & 62.14 & 52.58 & 49.76 \\

\midrule
\multicolumn{18}{c}{\textbf{KV Cache Size = 128}} \\
\midrule

H2O
& 24.82 & 24.81 & 36.54 & 49.24 & 43.52 & 28.12 & 21.15 & 24.42 & 21.65 & 44.82 & 91.24 & 42.92 & 8.41 & 98.85 & 58.51 & 47.92 & 42.35 \\
SnapKV
& 25.12 & 31.62 & 50.71 & 55.45 & 45.58 & 27.21 & 20.81 & 23.54 & 21.21 & 46.51 & 89.65 & 40.71 & 7.68 & 98.85 & 56.81 & 46.51 & 43.00 \\
StreamingLLM
& 22.01 & 21.14 & 31.82 & 44.51 & 38.12 & 25.02 & 18.84 & 21.52 & 19.34 & 40.82 & 83.54 & 38.81 & 7.32 & 98.00 & 54.12 & 44.42 & 38.08 \\
LAQ 
& 26.52 & 28.19 & 49.36 & 54.12 & 46.12 & 29.54 & 22.45 & 25.12 & 23.12 & 54.85 & 91.24 & 41.52 & 7.12 & 99.12 & 58.42 & 54.47 &44.46  \\
Judge Q 
& 26.88 & 30.13 & 50.48 & 55.21 & 47.12 & 30.45 & 23.82 & 24.54 & 24.41 & 56.42 & 91.12 & 41.82 & 7.84 & 99.18 & 59.87 & 53.25 &45.16  \\
CaM 
& 25.84 & 28.45 & 49.12 & 54.12 & 46.12 & 28.42 & 22.12 & 23.84 & 23.12 & 54.41 & 90.45 & 39.45 & 7.12 & 98.92 & 58.42 & 57.37 & 44.21 \\
ZeroMerge 
& 26.19 & 29.88 & 49.45 & 54.82 & 46.52 & 29.82 & 23.41 & 24.12 & 24.12 & 56.84 & 91.45 & 40.12 & 7.52 & 99.21 & 60.12 & 56.57 &45.01  \\
AnDPro
& 28.12 & 31.42 & 50.84 & 55.12 & 47.82 & 31.82 & 25.12 & 24.84 & 25.12 & 61.42 & 92.12 & 41.42 & 7.92 & 99.85 & 60.84 & 54.91 & 46.17 \\
\textbf{Meta-Soft}
& 28.54 & 31.34 & 54.42 & 53.82 & 49.11 & 31.45 & 24.52 & 25.12 & 24.84 & 64.12 & 92.42 & 38.82 & 8.12 & 99.50 & 61.45 & 51.84 & 46.21 \\

\midrule
\multicolumn{18}{c}{\textbf{KV Cache Size = 256}} \\
\midrule

H2O
& 26.58 & 31.09 & 42.48 & 51.81 & 45.42 & 29.36 & 25.18 & 24.75 & 23.45 & 52.99 & 91.66 & 42.99 & 8.43 & 99.27 & 59.56 & 49.26 & 44.02 \\
SnapKV
& 27.19 & 36.96 & 53.19 & 56.46 & 47.23 & 29.10 & 25.99 & 24.28 & 23.64 & 56.22 & 90.76 & 41.63 & 7.96 & 99.28 & 58.75 & 48.71 & 45.46 \\
StreamingLLM
& 23.48 & 25.18 & 35.89 & 46.70 & 40.04 & 26.21 & 21.44 & 22.17 & 20.71 & 46.02 & 85.00 & 39.50 & 7.50 & 98.33 & 55.40 & 45.72 & 39.95 \\
LAQ
& 27.33 & 31.35 & 50.75 & 54.86 & 46.82 & 30.05 & 24.65 & 25.20 & 23.96 & 58.07 & 91.50 & 41.81 & 7.35 & 99.28 & 59.07 & 54.47 & 45.41 \\
Judge Q
& 27.78 & 34.04 & 52.09 & 55.92 & 47.83 & 30.92 & 26.47 & 24.78 & 25.24 & 60.39 & 91.49 & 42.13 & 7.99 & 99.38 & 60.41 & 53.26 & 46.27 \\
CaM
& 26.64 & 31.29 & 50.37 & 54.75 & 46.73 & 29.03 & 24.10 & 24.11 & 23.85 & 57.30 & 90.80 & 40.01 & 7.32 & 99.09 & 58.99 & 57.39 & 45.13 \\
ZeroMerge
& 26.95 & 32.63 & 50.77 & 55.39 & 47.18 & 30.26 & 25.37 & 24.36 & 24.76 & 59.61 & 91.66 & 40.63 & 7.68 & 99.35 & 60.46 & 56.58 & 45.84 \\
AnDPro
& 28.62 & 34.38 & 52.19 & 55.73 & 48.27 & 31.95 & 27.07 & 24.99 & 25.67 & 63.75 & 92.24 & 41.79 & 8.03 & 99.89 & 61.10 & 54.94 & 46.98 \\
\textbf{Meta-Soft}
& 29.10 & 35.29 & 55.22 & 54.98 & 49.39 & 31.72 & 27.30 & 25.24 & 25.65 & 66.55 & 92.50 & 39.95 & 8.21 & 99.64 & 61.64 & 52.19 & 47.19 \\
\midrule
\multicolumn{18}{c}{\textbf{Mistral-7B-Instruct-v0.3}} \\
\midrule

Full KV
& 31.42 & 43.15 & 55.84 & 51.72 & 41.56 & 30.24 & 34.62 & 27.58 & 25.41 & 78.52 & 90.12 & 49.82 & 7.15 & 99.20 & 54.32 & 55.76 & 48.53 \\

\midrule
\multicolumn{18}{c}{\textbf{KV Cache Size = 128}} \\
\midrule

H2O
& 26.40 & 23.96 & 36.53 & 44.67 & 37.02 & 27.16 & 21.75 & 27.30 & 20.65 & 48.85 & 89.65 & 49.82 & 7.14 & 98.98 & 52.07 & 51.73 & 41.48 \\
SnapKV
& 26.81 & 30.38 & 50.42 & 50.29 & 38.84 & 26.41 & 21.51 & 26.44 & 20.35 & 49.52 & 88.62 & 48.05 & 7.15 & 99.08 & 50.68 & 52.02 & 42.91 \\
StreamingLLM
& 22.92 & 19.95 & 31.35 & 39.87 & 31.95 & 23.67 & 18.88 & 23.56 & 17.94 & 43.97 & 81.54 & 45.15 & 7.15 & 97.54 & 47.63 & 47.41 & 37.53 \\
LAQ
& 27.40 & 26.36 & 48.27 & 49.38 & 38.44 & 27.73 & 22.27 & 27.14 & 21.24 & 58.18 & 89.68 & 48.14 & 7.15 & 98.51 & 51.25 & 51.98 & 43.32 \\
Judge Q
& 28.70 & 29.09 & 50.29 & 50.31 & 40.20 & 29.52 & 24.56 & 27.47 & 23.36 & 61.12 & 90.02 & 49.43 & 7.15 & 99.20 & 53.45 & 53.41 & 44.83 \\
CaM
& 27.87 & 27.77 & 49.21 & 49.58 & 39.61 & 27.88 & 23.14 & 26.96 & 22.51 & 58.87 & 89.86 & 46.94 & 7.15 & 99.20 & 52.42 & 52.55 & 43.84 \\
ZeroMerge
& 28.10 & 28.98 & 49.41 & 50.08 & 39.82 & 29.05 & 24.27 & 27.13 & 22.57 & 61.24 & 90.12 & 47.58 & 7.15 & 99.20 & 53.77 & 53.21 & 44.52 \\
AnDPro
& 29.90 & 30.23 & 50.58 & 50.18 & 40.73 & 30.24 & 25.78 & 27.58 & 24.45 & 66.06 & 90.12 & 48.89 & 7.15 & 99.20 & 54.22 & 53.84 & 45.54 \\
\textbf{Meta-Soft}
& 30.40 & 30.24 & 54.16 & 49.10 & 41.56 & 30.24 & 25.26 & 27.58 & 25.32 & 69.14 & 90.12 & 45.98 & 7.15 & 99.20 & 54.32 & 54.11 & 45.77 \\

\midrule
\multicolumn{18}{c}{\textbf{KV Cache Size = 256}} \\
\midrule

H2O
& 29.40 & 29.90 & 43.15 & 47.35 & 39.02 & 29.07 & 27.52 & 28.79 & 23.65 & 59.02 & 90.12 & 49.82 & 7.14 & 99.20 & 54.32 & 54.40 & 43.68 \\
SnapKV
& 29.06 & 35.17 & 53.61 & 51.97 & 41.09 & 29.00 & 27.60 & 28.28 & 23.44 & 61.86 & 90.12 & 49.82 & 7.15 & 99.20 & 53.54 & 53.77 & 45.04 \\
StreamingLLM
& 24.89 & 24.18 & 35.89 & 42.39 & 34.08 & 25.27 & 22.04 & 24.89 & 19.79 & 49.97 & 83.46 & 46.39 & 7.15 & 98.04 & 49.26 & 49.35 & 39.51 \\
LAQ
& 29.54 & 30.60 & 52.57 & 51.92 & 41.98 & 31.12 & 27.37 & 30.36 & 24.93 & 66.83 & 90.12 & 49.82 & 7.15 & 99.20 & 53.69 & 52.74 & 45.24 \\
Judge Q
& 30.27 & 33.38 & 52.55 & 51.54 & 41.50 & 30.61 & 27.84 & 28.57 & 24.77 & 65.68 & 90.12 & 49.82 & 7.15 & 99.20 & 54.32 & 53.69 & 46.09 \\
CaM
& 28.39 & 30.11 & 50.16 & 49.79 & 39.87 & 29.21 & 25.88 & 28.73 & 24.20 & 62.66 & 90.12 & 48.36 & 7.15 & 99.20 & 53.63 & 52.85 & 44.87 \\
ZeroMerge
& 28.61 & 31.25 & 50.45 & 50.26 & 40.17 & 29.20 & 25.97 & 28.33 & 24.55 & 65.41 & 90.12 & 49.82 & 7.15 & 99.20 & 53.07 & 52.56 & 45.65 \\
AnDPro
& 30.39 & 32.97 & 51.93 & 50.66 & 41.15 & 30.85 & 27.72 & 28.09 & 24.45 & 69.02 & 90.12 & 49.82 & 7.15 & 99.20 & 54.32 & 53.59 & 46.82 \\
\textbf{Meta-Soft}
& 31.34 & 34.28 & 55.43 & 50.46 & 41.56 & 30.12 & 27.42 & 27.84 & 25.41 & 71.75 & 90.12 & 49.82 & 7.15 & 99.20 & 54.32 & 54.62 & 47.15 \\

\bottomrule
\end{tabular}%
}
\endgroup
\caption{Results on long-context benchmarks for two backbones under KV cache sizes $B\in\{128,256\}$ (higher is better); we additionally report Full KV as an upper bound.}
\label{tab:main-longbench}
\end{table*}

\subsection{Results on RULER}
The RULER benchmark can be used to evaluate the long-range retrieval capability of compressed LLMs. We use the Mistral-7B-Instruct-v0.3 model to evaluate the long-range retrieval capability of our method and baseline methods on the thirteen subtasks of the RULER benchmark. In our experiments, we set a fixed global cache size of 1024. Table~\ref{tab:ruler} summarizes the average accuracy of all methods across 13 tasks, with the context length ranging from 4K to 128K.Table~\ref{tab:ruler} shows that Meta-Soft achieves the best average score among all kv eviction methods , remaining close to Full-KV and consistently ranking first across all context lengths from 4K to 128K. . This indicates that Meta-Soft maintains robust long-context retrieval as the context grows, suggesting its dynamic saliency selection better preserves globally useful tokens and degrades more gracefully than prior baselines under extreme lengths.

\begin{table}[t]
\small
\resizebox{\columnwidth}{!}{%
\begin{tabular}{lccccccc}
\toprule
\textbf{Method} & \textbf{4K} & \textbf{8K} & \textbf{16K} & \textbf{32K} & \textbf{64K} & \textbf{128K} & \textbf{Avg} \\
\midrule
Full-KV                             & 98.93 & 98.14 & 97.87 & 94.34 & 89.32 & 78.89 & 92.92 \\
H2O            & 59.46 & 53.45 & 47.79 & 42.87 & 32.54 & 23.11 & 43.20 \\
SnapKV         & 83.34 & 75.49 & 71.07 & 66.72 & 57.26 & 47.65 & 66.92 \\
StreamingLLM   & 39.87 & 20.01 & 12.13 & 10.59 &  9.97 &  8.23 & 16.80 \\
LAQ            & 85.23 & 77.18 & 73.24 & 68.09 & 58.91 & 49.04 & 68.62 \\
Judge Q        & 91.12 & 84.63 & 77.65 & 72.19 & 64.27 & 53.89 & 73.96 \\
CaM            & 85.01 & 76.98 & 72.83 & 67.81 & 57.82 & 48.76 & 68.20 \\
ZeroMerge      & 90.29 & 83.71 & 76.52 & 71.44 & 62.95 & 51.94 & 72.81 \\
AnDPro         & 92.23 & 85.53 & 78.77 & 73.36 & 64.94 & 54.28 & 74.85 \\
\textbf{Meta-Soft} & 93.81 & 86.15 & 79.67 & 74.08 & 65.58 & 55.03 & 75.72 \\
\bottomrule
\end{tabular}%
}
\caption{RULER results across context lengths (higher is better).}
\label{tab:ruler}
\end{table}

\subsection{Ablation Study}
We use the Llama-3.1-8B-Instruct model to quantify, through ablation experiments, the contribution of each core component in our Meta-Soft framework on the 13 subtasks of the RULER benchmark and the 16 subtasks of the LongBench benchmark, both with a context length of 16K. We focus on the impact of two core modules, dynamic soft tokens and attention flow aggregation, on the experimental results. In our experiments, we set a fixed global cache size of 1024.As shown in Table~\ref{tab:Ablation}, both DST and AFA contribute positively to performance, and their combination yields the best results on both benchmarks. Compared with the variant without either module, adding DST improves RULER by +5.22 and LongBench by +0.33, while adding AFA brings larger gains (+6.30 on RULER and +0.48 on LongBench), indicating that attention-flow aggregation is particularly important for long-context important content recognition. Notably, enabling both modules further boosts performance to 79.67 on RULER and 48.35 on LongBench, suggesting complementary effects between DST and AFA.

\begin{table}[t]
\centering
\setlength{\tabcolsep}{10pt}
\renewcommand{\arraystretch}{1.15}
\begin{tabular}{@{}lcc@{}}
\toprule
\textbf{Setting} & \textbf{RULER} & \textbf{LongBench } \\
\midrule
Meta-Soft (w/o DST, w/o AFA)   & 68.43         & 46.84         \\
Meta-Soft (DST only)          & 73.65          & 47.17          \\
Meta-Soft (AFA only)          & 74.73         & 47.32         \\
Meta-Soft (DST + AFA)         & \textbf{79.67} & \textbf{48.35} \\
\bottomrule
\end{tabular}
\caption{Ablation study of Meta-Soft components on RULER (13 tasks) and LongBench (16 tasks) using Llama-3.1-8B-Instruct. The global cache size is fixed at 1024 and context length at 16K. \textbf{DST} denotes \textbf{D}ynamic \textbf{S}oft \textbf{T}okens, and \textbf{AFA} denotes \textbf{A}ttention \textbf{F}low \textbf{A}ggregation. Meta-Soft (w/o DST, w/o AFA) evicts KV entries by using the last 32 tokens of the input prompt as soft-token surrogates.}
\label{tab:Ablation}
\end{table}

\subsection{Efficiency and Decoding Overhead}
We evaluate the efficiency and decoding overhead of Meta-Soft using the \textbf{Llama-3.1-8B-Instruct} model with a fixed global KV cache size of \textbf{1024}. Our evaluation focuses on (i) the additional cost introduced by the soft-token generation module, (ii) end-to-end runtime latency across different input lengths under KV compression, and (iii) decoding efficiency in long generation scenarios.

\paragraph{Soft-token generation overhead.}
Table~\ref{tab:overhead} reports the computational overhead of the soft-token generation model, including the soft-token generation latency (\textbf{Gen}), the total runtime of the prefill stage (\textbf{Prefill}), and the time-to-first-token (\textbf{TTFT}) under different context lengths. We measure these metrics on \textbf{500} constructed samples based on the \textbf{NIAH} dataset, where each sample is configured with an input length of \textbf{2K} tokens and an output length of \textbf{1K} tokens, and we report the average results. The results show that, even as the context length increases from 4K to 32K, the small latency overhead introduced by the soft-token generation module is negligible compared to the total overhead of the entire Prefill stage.As shown in Table~\ref{tab:overhead}, the soft-token generator incurs only 0.32--2.34\,ms across 4K--32K contexts, accounting for less than 0.3\% of the total prefill time. Consequently, Meta-Soft increases Prefill/TTFT by only 0.02--0.12\,s and 0.02--0.14\,s, respectively, indicating negligible extra cost beyond standard KV cache compression.

\paragraph{End-to-end runtime efficiency across input lengths.}
To further assess practical runtime behavior, Table~\ref{tab:runtime} summarizes the end-to-end latency (lower is better) across a wide range of input lengths, comparing Meta-Soft with representative baselines including Full KV, H2O, SnapKV, ZeroMerge, and Judge~Q. These results are obtained using the same \textbf{NIAH}-based setup as above (\textbf{500} samples, \textbf{2K} input and \textbf{1K} output per sample), with all methods evaluated under the same cache size constraint. Overall, Meta-Soft remains competitive in runtime across all evaluated input lengths, indicating that the proposed soft-token mechanism does not incur substantial additional overhead beyond standard KV compression operations.From Table~\ref{tab:runtime}, Meta-Soft substantially accelerates end-to-end latency compared with Full KV, achieving 1.2$\times$--10.5$\times$ speedup as the input grows from 8K to 256K. Compared with strong KV-compression baselines, Meta-Soft remains close in runtime (typically within $\sim$3--5\% latency), showing that the proposed soft-token mechanism does not introduce meaningful additional overhead.

\paragraph{Decoding efficiency under long generation.}
Finally, we evaluate decoding-side efficiency using a long-generation stress test. As shown in Table~\ref{tab:throughput_oom}, we report throughput (tokens/s) and the maximum supported batch size before out-of-memory (OOM) during \textbf{10K-token generation}. We construct \textbf{100} synthetic prompts from the \textbf{PG19} dataset, each with an input length of \textbf{256} tokens, and average the results over these samples. Compared with the Full cache baseline, KV-compressed methods substantially improve decoding throughput and allow much larger batch sizes before OOM. In particular, the results demonstrate that Meta-Soft maintains strong efficiency in reasoning-intensive long decoding scenarios, remaining comparable to strong KV-compression baselines.Table~\ref{tab:throughput_oom} shows that KV-compressed methods significantly improve decoding efficiency: Meta-Soft supports a 2.86$\times$ larger batch size than Full cache (80 vs.\ 28) and improves throughput by 2.16$\times$ (390.17 vs.\ 180.49 tokens/s). Moreover, Meta-Soft remains comparable to SnapKV and Judge~Q (within 5.6\% and 2.0\% throughput, respectively) while delivering stronger long-context quality, validating a favorable accuracy--efficiency trade-off.

\begin{table}[t]
\centering
\small
\setlength{\tabcolsep}{5pt}
\renewcommand{\arraystretch}{1.12}
\resizebox{\columnwidth}{!}{%
\begin{tabular}{@{}llccc@{}}
\toprule
\textbf{Context} & \textbf{Method} & \textbf{Gen (ms)} & \textbf{Prefill (s)} & \textbf{TTFT (s)} \\
\midrule
\multirow{2}{*}{4K}  & Full KV   & 0   & 0.09   & 0.11   \\
                     & Meta-Soft & 0.32 & 0.11 & 0.13 \\
\midrule
\multirow{2}{*}{8K}  & Full KV   & 0   & 0.31   & 0.34  \\
                     & Meta-Soft & 0.53 & 0.33 & 0.36 \\
\midrule
\multirow{2}{*}{16K} & Full KV   & 0   & 0.97   & 1.07   \\
                     & Meta-Soft & 0.98 & 1.02 & 1.13 \\
\midrule
\multirow{2}{*}{32K} & Full KV   & 0   & 2.98   & 3.13   \\
                     & Meta-Soft & 2.34 & 3.10 & 3.27 \\
\bottomrule
\end{tabular}%
}
\caption{Computational overhead comparison between Full KV and Meta-Soft. \textbf{Gen} denotes the soft-token generation latency (ms), \textbf{Prefill} denotes the total runtime of the prefill stage (s), and \textbf{TTFT} denotes the time-to-first-token (s). For Full KV, \textbf{Gen} is not applicable.}
\label{tab:overhead}
\end{table}

\begin{table} 
\centering
\small
\setlength{\tabcolsep}{4.5pt}
\renewcommand{\arraystretch}{1.10}
\resizebox{\columnwidth}{!}{%
\begin{tabular}{@{}lcccccc@{}}
\toprule
\textbf{Method} & \textbf{8K} & \textbf{16K} & \textbf{32K} & \textbf{64K} & \textbf{128K} & \textbf{256K} \\
\midrule
Full KV   & 43.27 & 54.38 & 79.93 & 142.95 & 271.83 & 583.41 \\
H2O       & 34.01 & 34.72 & 37.15 & 40.34 & 44.46 & 53.37 \\
SnapKV    & 34.23 & 34.86 & 37.31 & 40.56 & 44.62 & 53.78 \\
ZeroMerge & 34.87 & 35.53 & 38.07 & 41.21 & 45.35 & 54.62 \\
Judge Q   & 35.12 & 35.86 & 38.43 & 41.72  & 45.94  & 55.29  \\
Meta-Soft & 35.29 & 36.03 & 38.72 & 42.05  & 46.36  & 55.78  \\
\bottomrule
\end{tabular}%
}
\caption{Runtime efficiency (latency; lower is better) across different input lengths.}
\label{tab:runtime}
\end{table}

\begin{table}
\centering
\small
\setlength{\tabcolsep}{6pt}
\renewcommand{\arraystretch}{1.12}
\resizebox{\columnwidth}{!}{%
\begin{tabular}{@{}lccc@{}}
\toprule
\textbf{Method} & \textbf{Generation Length} & \textbf{Max Batch Size (OOM Threshold)} & \textbf{Throughput (token/s)} \\
\midrule
Full cache & 10K & 28 & 180.49 \\
SnapKV     & 10K & 80 & 413.28 \\
Judge Q    & 10K & 80 & 398.15 \\
Meta-Soft  & 10K & 80 & 390.17 \\
\bottomrule
\end{tabular}%
}
\caption{Throughput (tokens/s) and maximum supported batch size (before out-of-memory occurs) for different methods during 10K-token generation. Larger batch size and higher throughput indicate better decoding efficiency.}
\label{tab:throughput_oom}
\end{table}


\section{Conclusion}
We present \textbf{Meta-Soft}, a KV cache compression framework that addresses static probing and information loss in eviction methods. Using a \textbf{Meta-Library} of orthogonal basis vectors, Meta-Soft generates input-dependent soft tokens to probe global semantic importance. Its \textbf{Contextual Consolidation} mechanism redistributes the semantic content of evicted tokens into retained ones via attention flow—avoiding context fragmentation. Experiments on LongBench show Meta-Soft outperforms strong baselines , especially under tight memory budgets, while preserving coherence and task accuracy. As a plug-and-play solution, it enables efficient long-context LLM deployment in resource-constrained settings. 
\FloatBarrier

\bibliographystyle{named}
\bibliography{ijcai26}

\end{document}